# Application of machine learning for predicting the spread of COVID-19

Xiaoxu Zhong[*], Yukun Ye

*School of Mechanical Engineering, Purdue University, West Lafayette, IN 47906, United States*

## Abstract

The spread of diseases has been studied for many years, but it receives a particular focus recently due to the outbreak and spread of COVID-19. Studies show that the spread of COVID-19 can be characterized by the Susceptible-Infectious-Recovered-Deceased (SIRD) model with containment coefficients (due to quarantine and keeping social distance). This project aims to apply the machine learning technique to predict the severity of COVID-19 and the effect of quarantine, keeping social distance, working from home, and wearing masks on the transmission of disease. This work can deepen our understanding of the COVID-19 transmission and reveals the importance of following policies.

## Introduction

The outbreak of COVID-19 has brought a significant impact on the world. More than forty million people have been infected and millions of people have died. Especially, governments publish policies that people should stay at home if they feel uncomfortable; people should follow strict measures including keeping social distance and wearing masks in public; and people should work from home if possible [1, 2]. Extended periods of lockdowns are also harmful for the economy of any country [3, 4]. Confirmed case numbers in many cities show that the growth is exponential at the beginning, but the growth becomes sub-exponential after people follow the policies, such as the quarantine [5]. Thus, the spread of COVID-19 decreases due to the effective containment.

The spread of COVID-19 can be well described by the SIRD-X model, which adds two "containment" coefficients (corresponding to quarantine and keeping social distance) to the popular Susceptible-Infectious-Recovered-Deceased (SIRD) model [5, 6, 7]. The SIRD-X model reads:

$$\partial_t S = -\alpha SI - \kappa_0 S,$$
$$\partial_t I = \alpha SI - (\beta + \mu)I - \kappa_0 I - \kappa I,$$
$$\partial_t R = \beta I,$$
$$\partial_t D = \mu I,$$
$$\partial_t X = \kappa_0 S + (\kappa + \kappa_0)I,$$

where $S, I, R, D, X$ are the number of susceptible, currently infected (test results are positive), recovered (test results are negative but they were infected previously), death, and those who are alive but do not take part in transmission (due to quarantine), respectively. $\alpha, \beta, \mu, \kappa, \kappa_0$ are the rates of infection, recovery, mortality, containment that only affects symptomatic infected due to quarantine, and the containment that is effective in the whole population due to work from home and curfew, respectively.

In this project, we apply the machine learning technique for the SIRD-X model. We focus on the total number of deaths $D$ and the maximum number of infected individuals $\max\{I(t)\}$, because $D$ reflects the mortality of the disease, and $\max\{I(t)\}$ indicates whether the situation is under control with the current medical system. Least-square regression, Bayesian linear regression [8], support vector regression [9, 10], ANN, and other approaches are commonly used for regression. Here, we use the neural network to predict $D$ and $\max\{I(t)\}$. Besides regression, we are usually only interested in whether the number of deaths and infected individuals is below the threshold or not. For instance, if the maximum number of infected individuals is larger than 20% of the whole population, our current medical system is likely to break down because the available empty beds in the hospital are far less than the needs, such as the situation in New York in March and April in 2020. Thus, it is also meaningful to apply the support vector machine [11] or logistic regression techniques for the study of the spread of COVID-19.

This project is organized as follows. At first, we generate data by solving the SIRD-X model. The coefficients $\alpha, \beta, \mu, \kappa, \kappa_0$ are randomly picked in reasonable space. Next, we classify the disease to have high mortality if more than 2% of the population is dead due to infection. Also, we classify the situation to be out of control, and thus, the current medical system needs upgrading if more than 10% of the population showing similar symptoms at the same time. Next, we apply the deep neural network for regression, and we apply the SVM for classification. At last, we analyze the effects of wearing masks and keeping social distance (reducing the rate of infection $\alpha$), quarantine (reducing $\kappa$, i.e., active activity interacted with infected individuals) and working from home (reducing $\kappa_0$, i.e., active activity of the whole population), on the spread of COVID-19.

## Methodology

Our workflow is shown in Figure 1. At first, we collect data by solving the SIRD-X model. Second, we apply the deep neural network to predict $D$ and $\max\{I(t)\}$. Third, we apply the logistic regression and support vector machine (SVM) for classification. After training and validating our regression and classification models, we perform sobol sensitivity analysis to measure the importance of the rate of infection $\alpha$, the containment coefficient $\kappa$ which is relevant with quarantine, and the containment coefficient $\kappa_0$ which is relevant with work from home. At last, we answer the question that are the policies effective to reduce the spread of COVID-19 or not.

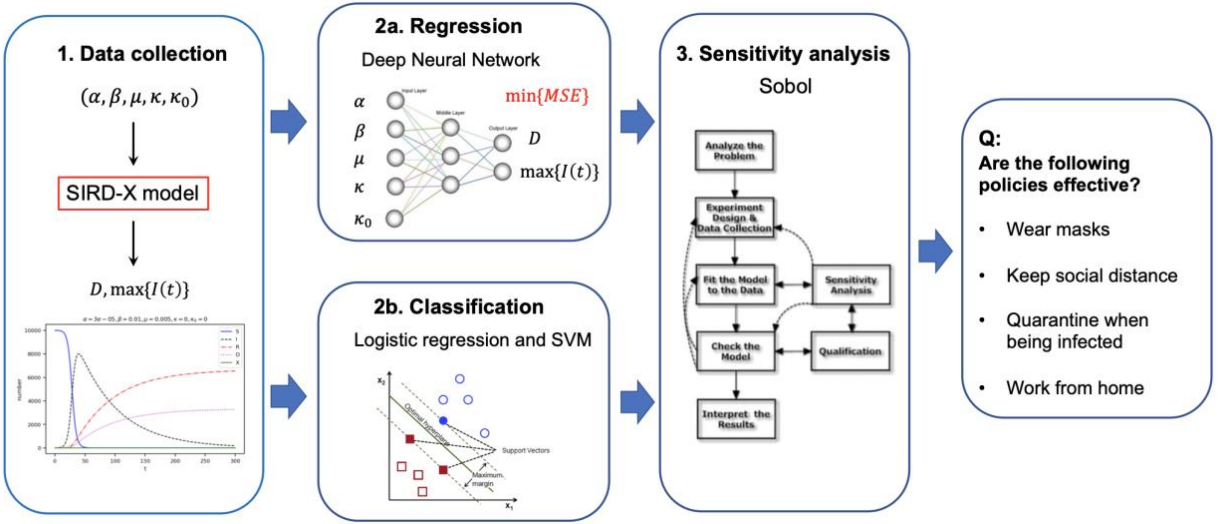

**Fig 1. The workflow**

### 1. Data collection

We assume $\alpha, \beta, \mu, \kappa, \kappa_0$ to be constants and $\alpha \in [10^{-6}, 3 \times 10^{-5}]$, $\beta \in [0.01, 0.1]$, $\mu \in [0.001, 0.005]$, $\kappa \in [0, 0.02]$, $\kappa_0 \in [0, 0.005]$. The total number of individuals is assumed to be 10,000, i.e., $N = S + I + R + D + X = 10000$. Additionally, we assume that 3 individuals are infected at $t = 0$, i.e., $S(0) = 9997, I(0) = 3, R(0) = D(0) = X(0) = 0$.

We randomly pick 1000 samples $[\alpha, \beta, \mu, \kappa, \kappa_0]$, and we use 1st-order explicit Euler method to solve the SIRD-X model to obtain the total number of deaths $D$ and the maximum number of infected individuals $\max\{I(t)\}$.

### 2. Regression

We use artificial neural network to predict the total number of deaths and the maximum number of infected individuals. The inputs and outputs are scaled to [0, 1]. We have five neurons in the input layer and two neurons in the output layer. The hyper-parameters include the number of hidden layers, the number of neurons in each hidden layer, the activation function, the optimizer, etc [12]. For the sake of simplicity, we only perform the sensitivity study to determine the optimal number of hidden layers and the activation function. Other hyper-parameters are fixed in this project. We use 32 neurons in each hidden layer, and we use the mean-squared-error as the loss function. Also, the optimizer is selected as "adam" in all cases. Additionally, we use the early-stop method with patience = 50 to avoid the overfitting.

### 3. Classification

We create a data driven machine learning model to predict the total number of deaths (D) and the maximum number of infected individuals max{$I(t)$} regarding to $\alpha, \beta, \mu, \kappa, \kappa_0$, respectively. the total number of deaths (D) and the maximum number of infected individuals max{$I(t)$} is transformed into a binary variable (0 ,1) for the classification task. We determine the disease have high mortality rate if more than 2% individuals are dead, i.e., D>200, then the output variable (D) is set to 1. if the maximum number of individuals, whose test results are positive at the same time, is less than 10%, i.e., ml < 1000, we classify this situation to be well controlled, then the output variable max{$I(t)$} is set to 0. We develop the SVM model and logistic regression model by using all the input variables and binary death number (D) and current infected numbers (ml) as the output variable to calculate the accuracy and generate the confusion matrix.

## Results

### 1. Regression

Fig. 2 shows the distribution of the number of death populations, $D$, and the maximum number of infected populations, max {$I(t)$}. The average of $D$ and max {$I(t)$} are 254 and 1541, respectively. The architecture of artificial neural network is shown in Fig 1 (2a). We also add the user-bias to the input layer and the hidden layers. Here, we perform the sensitivity study to determine the optimal number of hidden layers and the activation function. However, the $R^2$-score varies at each run which brings difficulty in deciding which value of the hyper-parameter is optimal. To overcome this difficulty, we use 10-Fold cross-validation in evaluation such that our results are robust. The impact of the number of hidden layers on the performance of the ANN model is listed in Table 1. Using 2 hidden layers or 4 hidden layers yield best results. Considering that using more hidden layers is more likely to overfit the data, we use 2 hidden layers in the subsequent studies. The impact of the activation function on the performance of the ANN model is listed in Table 2. The four activation functions yield similar results with $R^2 \approx 0.997$. Thus, the activation function has little influence of our results, and we use "Relu" as the action function in subsequent studies. We use "early stop" with patience 50 to prevent overfitting.

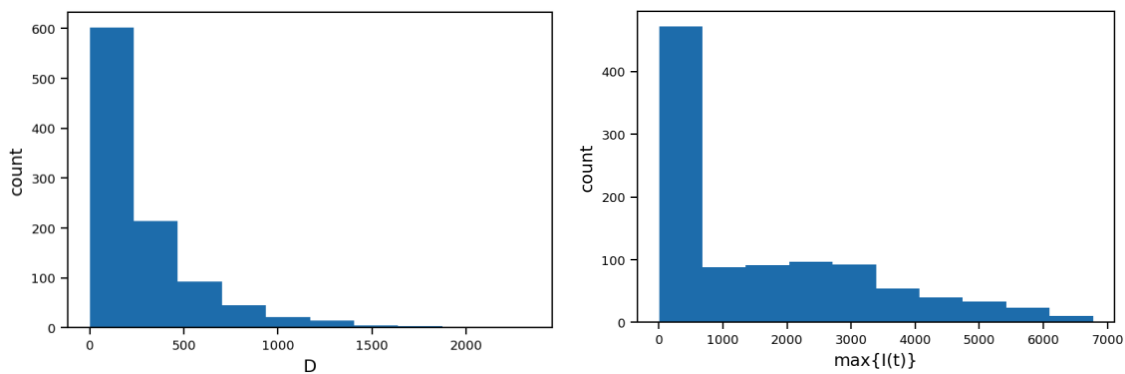

**Fig 2. The histogram of $D$ and max{$I(t)$}.**

**Table 1. The impact of the number of hidden layers on the performance of ANN model.**

| The number of hidden layers | Average $R^2$ score | Standard deviation of $R^2$ score |
|---|---|---|
| 1 | 0.9877 | 0.0137 |
| 2 | 0.9966 | 0.0057 |
| 3 | 0.9931 | 0.0149 |
| 4 | 0.9962 | 0.0058 |

**Table 2. The impact of the activation function on the performance of ANN model.**

| The activation function | Average $R^2$ score | Standard deviation of $R^2$ score |
|---|---|---|
| Sigmoid | 0.9968 | 0.0045 |
| Tanh | 0.9956 | 0.0043 |
| Relu | 0.9970 | 0.0051 |
| Leaky_relu | 0.9968 | 0.0060 |

After deciding the optimal number of hidden layers and the activation function, we split the data into 80% training data and 20% test data. Fig. 3 shows the comparisons of the (training and testing) data and the model prediction. Our model agrees well with the data which illustrates the validity of our ANN model. Additionally, the training loss and testing loss are shown in Fig. 4. It is found that the mean-squared-error is on the magnitude of $10^{-5}$.

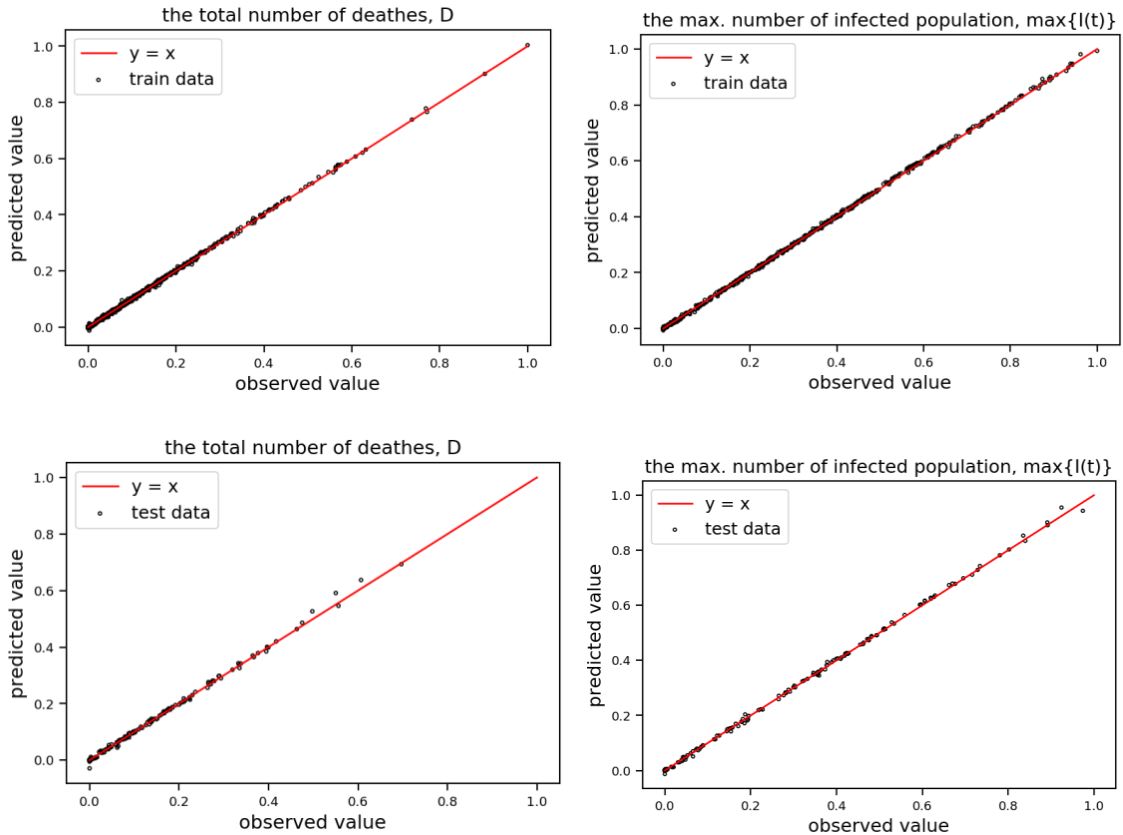

**Fig 3. The comparisons of data and the model prediction**

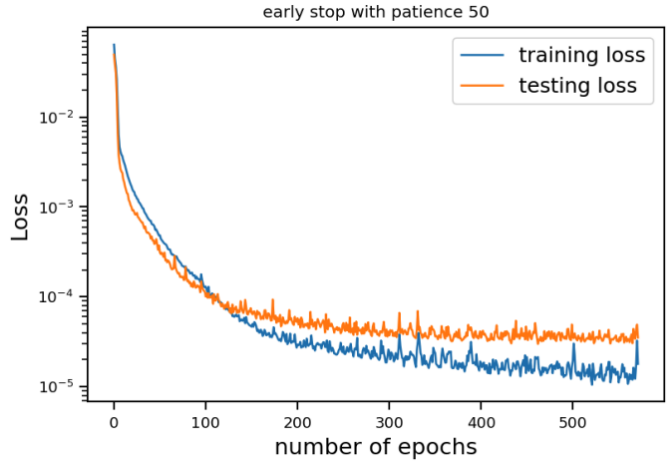

**Fig 4. The evolution of the training and testing loss.**

## 2. Classification

Fig.5 shows the confusion matrices which generated by the logistic regression model, Fig.6 shows the confusion matrices generated by the support vector machine (linear, polynomial and radial basis functions). Table 3 compares the accuracy between different models. We can see the accuracy for the total number of deaths is 1, which means the test data is highly close to the predict data; the accuracy of and the maximum number of infected populations, $\max\{I(t)\}$ is 0.925, which means 92.5% of test data meet the predict data. And we observe no matter which ways we use to calculate the accuracy, the results remain same.

**Table 3. Accuracy of D & $\max\{I(t)\}$ in different models.**

| Cases | Accuracy in Logistic regression model | Accuracy in SVM (linear kernel) | Accuracy in SVM (Polynomial Kernel) | Accuracy in SVM (RBF kernel) |
|---|---|---|---|---|
| D | 1.000 | 1.000 | 1.000 | 1.000 |
| $\max\{I(t)\}$. | 0.925 | 0.925 | 0.925 | 0.925 |

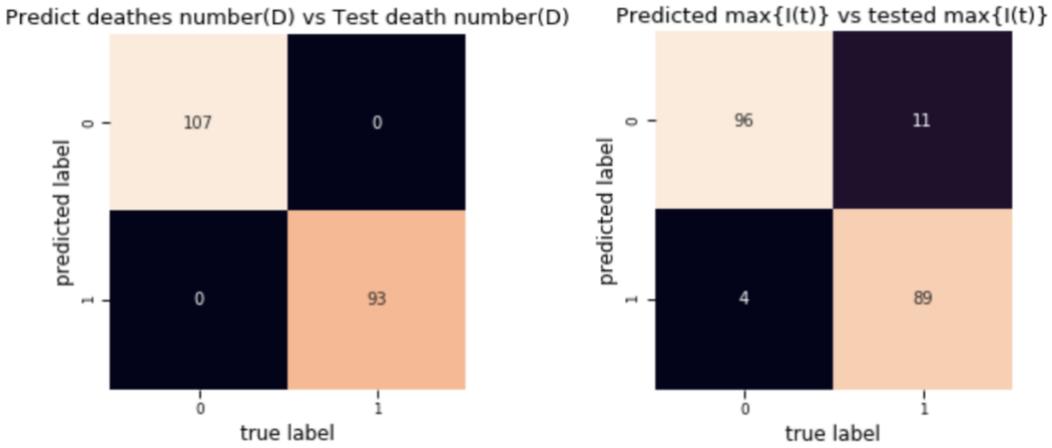

**Fig 5. The confusion matrix generated by logistic regression model**

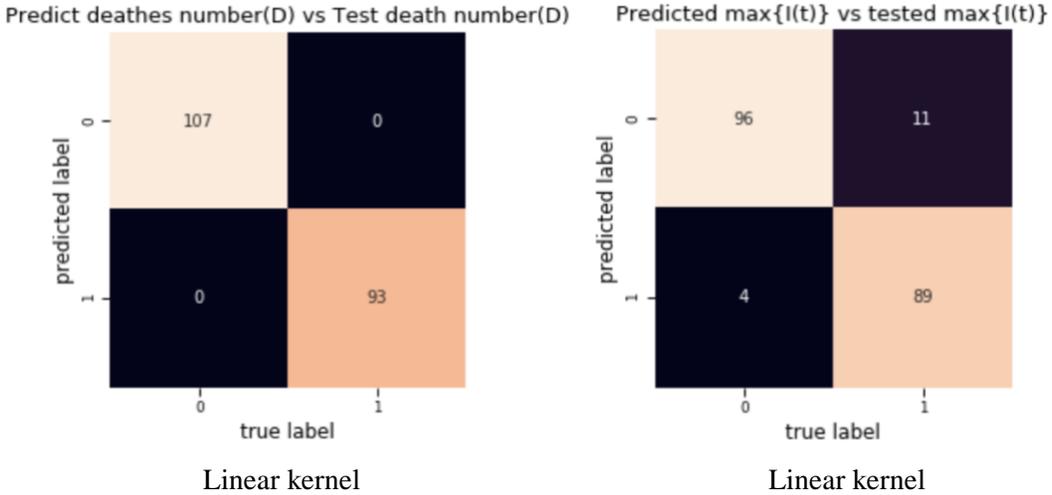

| Linear kernel | Linear kernel |

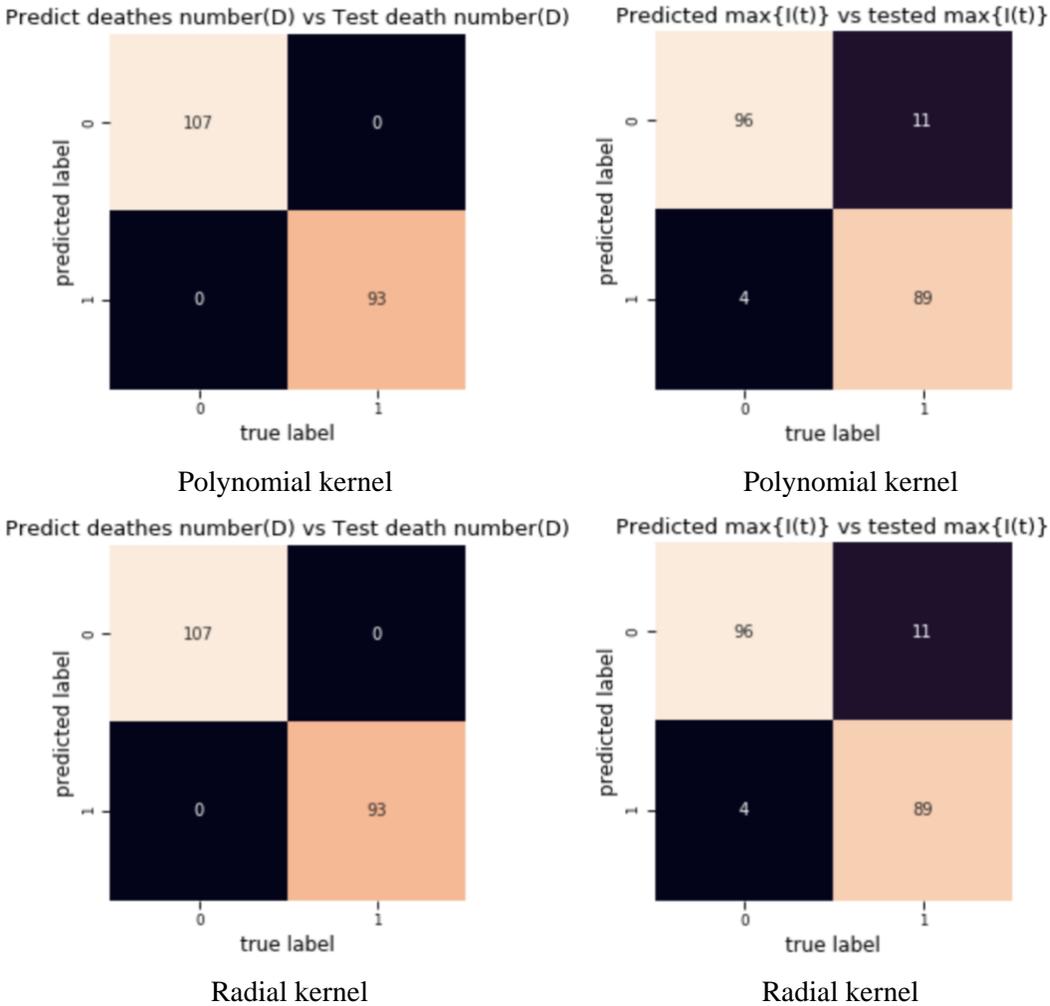

| | |
|---|---|
| Polynomial kernel | Polynomial kernel |
| Radial kernel | Radial kernel |

**Fig 6. The confusion matrix generated by SVM**

### 3. Sensitivity analysis

Here, we use the sobol sensitivity analysis [13] to measure the importance of each input parameters to the outputs, as shown in Fig. 7. The rate of infection $\alpha$ and the rate of recovery $\beta$ have more significant impacts on the number of death $D$ and infected individuals max $\{I(t)\}$ than the rate of mortality $\mu$ and containment coefficients $\kappa_0$ and $\kappa$. This indicates that wearing masks which reduces the rate of infection is more effective than quarantine. Also, the government's policies should pay more attention to reducing the rate of infection and increasing the rate of recovery.

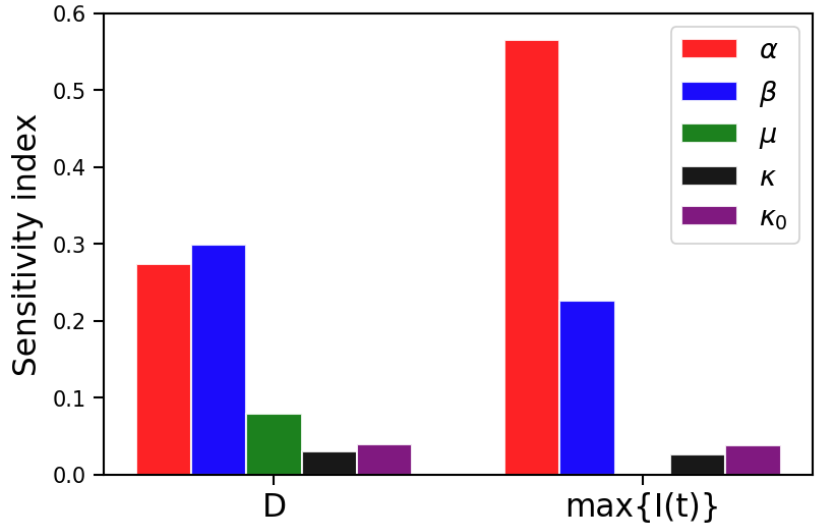

**Fig. 7. Sobol sensitivity analysis**

## 4. Quantitative analysis

To better understand the SIRD-X model, we manipulate the values of α, β, μ, κ, $κ_0$ to study the change of outputs. For instance, in the case of $α = 3 \times 10^{-5}, β = 0.01, μ = 0.005$, $κ = κ_0 = 0$, we obtain D = 6552 and max{I(t)} = 8002, as shown in Fig. 8. However, if we follow quarantine policies (κ = 0.02) and some of us work from home ($κ_0 = 0.005$), then the number of deaths D reduces from 6552 to 1053, and the maximum number of infected individuals max{I(t)} reduces from 8002 to 4752. Thus, the measures, such as obeying the quarantine policies when get infected, and working from home if possible, are also important.

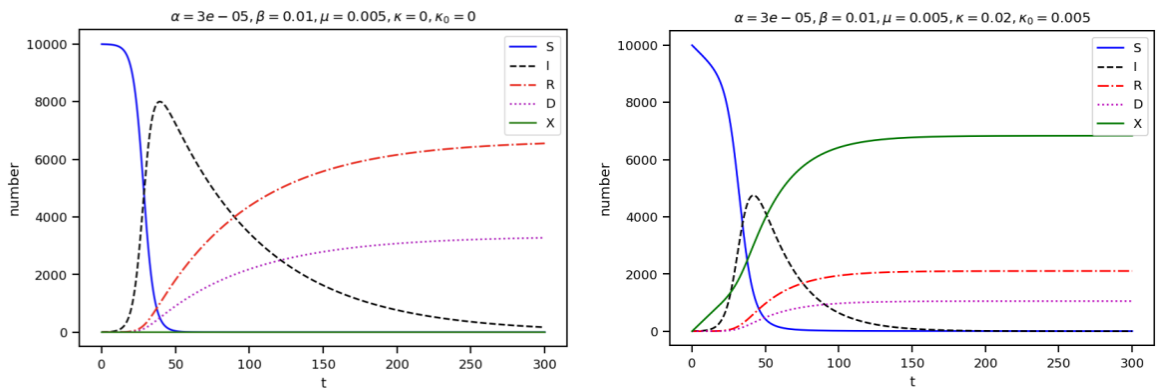

**Fig. 8. Quantitative analysis**

# Conclusions

Studies [5] show that the SIRD-X model can well predict the spread of COVID-19. Based on the data generated by the SIRD-X model, we propose a machine learning technique to predict the spread of COVID-19 in this project. We apply the neural network to predict the values of the maximum number of deaths and infected individuals. We perform the sensitivity analysis with cross-validation to determine the optimal number of hidden layers and the activation function. The value of $R^2$ reaches 0.997. Also, we find that both the logistic regression method and SVM can well predict the SIRD-X model. Thus, government can use our model to predict whether the virus has high mortality or not, and whether the beds in hospital are enough or not.

In addition, the results of the sobol sensitivity analysis indicate that the rate of infection and the rate of recovery are more important than the rate of mortality and containment coefficients. Thus, wearing masks is more effective than quarantine and working from home. Here, it is worth mentioning that the sensitivity analysis results do not mean that quarantine and working from home are useless, since the number of deaths and infected individuals reduce with increasing the containment coefficients $\kappa, \kappa_0$, as shown in Fig. 8.

Based on our results, we suggest everyone should wear masks in public, because it is the most effective way to reduce the spread of COVID-19. Government should pay more attention to reduce the rate of infection and increase the rate of recovery. In addition, quarantining when get infected, and working from home if possible, are also important measures to reduce the spread of COVID-19.

# Assumptions

- Policies such as quarantine, curfew, and work from home remove individuals from the interaction dynamics.
- $\alpha, \beta, \mu, \kappa, \kappa_0$ are constants.
- We assume $\alpha \in [10^{-6}, 10^{-4}]$, $\beta \in [0.01, 0.1]$, $\mu \in [0.001, 0.004]$, $\kappa \in [0, 0.01]$, $\kappa_0 \in [0, 0.01]$;

# Relevance to society

Deepen people's understanding on the outbreak and spread of disease. Reveal the significance of taking effective measures such as wearing masks, taking quarantine, and keeping social distance, on protecting the public health.